\pdfoutput=1

\documentclass[11pt]{article}

\usepackage{acl}

\usepackage{times}
\usepackage{latexsym}
\usepackage{tabularx}
\usepackage{graphicx}
\usepackage{booktabs}
\usepackage[T1]{fontenc}

\usepackage[utf8]{inputenc}

\usepackage{microtype}

\title{AraLegal-BERT: A pretrained language model for Arabic Legal text}

\author{Muhammad AL-Qurishi, Sarah AlQaseemi \and Riad Soussi  \\
	Elm Company, \\ Research Department,  \\ Riyadh 12382, Saudia Arabia}

\begin{document}
	\maketitle
	\begin{abstract}
		The effectiveness of the BERT model on multiple linguistic tasks has been well documented. On the other hand, its potentials for narrow and specific domains such as Legal, have not been fully explored. In this paper, we examine how BERT can be used in the Arabic legal domain and try customizing this language model for several downstream tasks using several different domain-relevant training and testing datasets to train BERT from scratch. We introduce the AraLegal-BERT, a bidirectional encoder Transformer-based model that have been thoroughly tested and carefully optimized with the goal to amplify the impact of NLP-driven solution concerning jurisprudence, legal documents, and legal practice. We fine-tuned AraLegal-BERT and evaluated it against three BERT variations for Arabic language in three natural languages understanding (NLU) tasks. The results show that the base version of AraLegal-BERT achieve better accuracy than the general and original BERT over the Legal text.
	\end{abstract}
	
	\section{Introduction}
	
	The impressive performance of Bidirectional Encoder Representations from Transformers (BERT)~\cite{devlin2018bert} inspired numerous authors to try and improve on the original BERT. This type of follow-up research goes in several different directions, including the development of specific solutions for various thematic domains. This is necessary because the vocabulary used in some fields significantly deviates from the language used for everyday purposes and may include specific meanings of certain phrases or atypical relationships between contextual elements. This problem can be partially resolved with domain-specific adjustments of the training process – a good example of this approach is demonstrated in the work by~\cite{chalkidis2020legal}, who created Legal-Bert specifically for mining text related to English legal text, improving the output of the standard Transformer algorithm within this domain. Another successful adaptation of the BERT concept was performed by~\cite{beltagy2019scibert,lee2020biobert}, who created models that were pre-trained on a compilation of scientific and Biomedical data from various fields, achieving significantly better performance on scientifically-related NLP tasks. Those examples show that BERT is far from a finished model at the moment, and that its current effectiveness could be enhanced even further, at least when it comes to relatively narrowly defined tasks.
	
	For Arabic language, several BERT-based models have consistently demonstrated superior performance on numerous linguistic tasks that require semantic understanding, outperforming all benchmarks on public datasets such as Arabic NER corpus (ANERcorp) or Arabic Reading Comprehension Dataset (ARCD), such as the works presented by~\cite{antoun2020arabert, abdul2020arbert} and mBERT by Google~\cite{devlin2018bert}.  This is largely a consequent of efficient transfer learning inherent in this model, which comes at a high computational price as this approach requires huge collections of training examples, followed by fine-tuning for specific downstream tasks. One great advantage of BERT is that the training phase can be skipped since it’s possible to take already pre-trained version of the model and train it further. However, it has been shown by~\cite{chalkidis2020legal,beltagy2019scibert,lee2020biobert} that generalist approach to pre-training doesn’t work well when BERT needs to be used within a domain with highly specific terminology, such as legal, science or medicine. There are two possible responses to this issue, namely, to continue specializing an already pre-trained model or to start training it from the beginning with relevant material from the domain.
	
	In this paper, we build a language model from scratch based on the original Bert~\cite{devlin2018bert}, this model is specific to Arabic legal texts with the aim of improving the performance of state-of-the-art in most of the tasks in language understanding and processing, especially Arabic legal texts. We believe that it is important to take into account the specific nature of legal documents and terminology, which affects the way sentences and paragraphs are constructed in this field. In fact, the extent of formal and semantic differences is such that some authors describe the linguistic content used for legal matters as almost a language of its own~\cite{chalkidis2020legal}. 
	
	By focusing on one Arabic legal text area, this paper attempts to reveal how NLP model can be adjusted to fit any thematic domain. Based on our experiments, we can confirm that pre-training BERT with examples from the Arabic legal domain from scratch provides a better foundation for working with documents containing Arabic legal vocabulary than using vanilla version of the algorithm. 
	
	We introduce the AraLegal-BERT, a Transformer-based model that have been thoroughly tested and carefully optimized with the goal to amplify the impact of NLP-driven solution concerning jurisprudence, legal documents, and legal practice. We fine-tuned AraLegal-BERT and evaluated it against three BERT variations for Arabic language in three NLU tasks. The results shows that the base version of AraLegal-BERT achieves better accuracy than the general and original BERT over the Legal text. AraLegal-BERT is a particularly efficient model that can match the output of computationally more demanding models while reaching its findings faster and expending far less resources. Consequently, it was observed that the base version of the model can reach comparable accuracy levels to larger versions of the large general and orginal BERT when they are trained with domain-relevant examples similar to ones use for testing our model. 
	
	\section{Related work}
	In a very short time, Transformers~\cite{vaswani2017attention} architecture has become the golden standard of machine learning methodologies in the field of linguistics~\cite{wolf2020transformers, su2022comprehensive}. Unparalleled success of BERT in combination with its flexibility lead to proliferation of tools based on it, created with a more narrowly defined vocabulary~\cite{young2018recent, brown2020language}. AraBERT~\cite{antoun2020arabert,abdul2020arbert},  are examples of such specialization, and given the number of Arabic speakers worldwide it could have considerable practical value. Since the model is trained for some of the most commonly used NLP tasks and proven effective across regional variations in morphology and syntax, those language models have a chance to become the default tool for analyzing Arabic text. Pre-training and fine tuning procedures that were described in this work may not be the most optimal ones, but the output of the localized model clearly indicated that the initial approach was correct. With further refinement, the model could become reliable enough for a wide range of real world applications.
	On the other hand, these models are based on data most of which were collected from the Modern Standard Arabic, and these language models may fail when the language changes to colloquial dialects~\cite{abdelali2021pre}. In addition, the performance of these language models may be affected when dealing with a language for a specific domain with special terms such as scientific, medical and legal terms~\cite{yeung2019effects}.
	
	The majority of BERT domain-specific forms are related to scientific or medical texts, as well as legal but all are in English language~\cite{beltagy2019scibert,lee2020biobert, chalkidis2020legal}. In a study by~\cite{alsentzer2019publicly} the main area of interest was clinical practice, so the authors developed two different variations by pre-training basic BERT and BIOBERT with examples from this domain, with positive results in both cases. Another interesting project of this kind was conducted by~\cite{beltagy2019scibert}, resulting in the creation of SCIBERT, a whole branch of variations optimized for use with scientific documents. In this case, two different optimization strategies were tested, including additional training and training from scratch using documents that include scientific terminology, with both approaches creating measurable improvements. A study by~\cite{chalkidis2020legal} included pre-training of Transformer models for English legal text with comparing three possible approaches for adaptation of BERT to thematic content niches – 1) using the vanilla model without any modifications; 2) introducing pre-training with datasets that contain examples from the target domain on top of standard training; and 3) Using only domain-relevant samples to train BERT from scratch. 
	
	Practically all of the BERT adaptations that include fine-tuning use the same approach for the selection of hyper-parameters as outlined in the original BERT formulation without even questioning it. Another research gap is observed when it comes to the possibility of using shallower models for executing domain-specific tasks. It could be argued that the impressive generalization capacity characteristic for deep models with lots of layers is wasted when the model works within a narrowly defined field, where linguistic rules are more streamlined and the vocabulary volume is more limited. Despite the fact that BERT has been the most successful deep learning model for many tasks related to the legal sphere, so far there have been no publicized attempts to develop a variation unique for this type of content especially in Arabic, which was largely a motivation for this study. Therefore, to the best of our knowledge, this is the first work to build a BERT-based language model for legal texts in Arabic.
	
	\section{AraLegal-BERT: Transformer Model Pre-trained with Arabic Legal Text}
	In order to be able to optimize BERT for working with Arabic legal documents, we followed the same procedures in the original BERT model~\cite{devlin2018bert}; however, for Arabic language we followed the same procedure in AraBERT~\cite{antoun2020arabert}.
	\subsection {Dataset} 
	Due to relative scarcity of publically available large-scale resources in Arabic lagal text, the training dataset had to be manually collected from several sources and included many regional variations. All the collected documents are in Arabic language related to several different sub-fields of legal practice, such as legislative documents, judicial documents, contracts and legal agreements, Islamic rules and Fiqh. All data was collected from public sources, and the final size of the dataset was 4.5GB. The final size of the training set after the removal of duplicates was approximately 13.7 million sentences. Table~\ref{dtab} describes the used dataset.
	
	\begin{table*}[!ht]
		\caption{The dataset used to train AraLegal-BERT}
		\resizebox{\textwidth}{!}{%
			\begin{tabular}{lll}
				\hline
				Type             & Sample Size & Desc                                                                             \\ \hline
				Books            & 6K          & Master and   PhD theses, research papers, magazines, dictionaries and Fiqh books \\
				Cases & 336K & Legal Cases in   KSA and Gulf countries which consists of copy rights, design rights, facts and appealing \\
				Terms and   laws & 3K          & Laws and   regulations in KSA and Gulf countries                                 \\
				others           & 5K          & Reports and studies, academic courses, forms, reports, contracts       \\ \hline
			\end{tabular}%
		}
		\label{dtab}
	\end{table*}
	
	\subsection {AraLegal-BERT} This version of the model was created by following up on the original pre-training with additional steps involving textual material from the Arabic legal domain. The authors of the original BERT indicated that 100,000 steps would be sufficient, but in our implementation the model was trained with up to half million steps to determine how extended pre-training with narrowly focused data samples affects performance on various linguistic tasks. The pre-training of BERT base~\cite{devlin2018bert} with general content involves a lot more steps, which is why the model tends to be most proficient with a vocabulary containing around 30,000 words commonly found in everyday language. It was presumed that with extended training using domain-focused examples this tendency can be partially reversed, with a positive impact on model accuracy. 
	
	Before we start the training, data pre-processing was performed and in this part we followed the same procedure in~\cite{antoun2020arabert}. To account for the unique nature of Arabic prefixes, sub-word segmentation was performed to separate all tokens into prefixes, stems, and suffixes as explained in~\cite{abdelali2016farasa}. This resulted in a vocabulary of around 64K words, which was used to pre-train the model and create AraLegal-BERT. Then we traind our model on masked language modeling (MLM) task where 15\% of the entire input sequence were used as tokens as 80\% of them were masked, 10 percent were replaced with a random token, and only 10\% were left in their natural state. This procedure allows the algorithms to derive conclusions based on entire words rather than just linguistic elements, which is better suited for the Arabic Language.  
	
	\section{Experimental procedure}
	\subsection{Pre-training stage} AraLegal-BERT was trained for approximately 50 epochs, involving a total of half million steps, which is similar to the original BERT pre-training procedure. We have trained our model in Elm research center using NVIDIA DGX1 with 8 GPUs. Batch size was set at 8 per gpu, so the total train batch size (w. parallel, distributed \& accumulation) is 512. The maximum sequence length is 512 tokens and a learning rate ranging from $1e-5$ to $5e-5$.
	
	\subsection{Fine-tuning }
	The authors of BERT~\cite{devlin2018bert} proposed an approach for determining the most optimal parameters for fine-tuning that is based on a search within a limited range. In this concept, the learning rate, length of training, size of training batch and dropout rate are all either fixed or can take one of a few possible discreet values. While no particular reason has been given for this approach, it has been widely replicated in studies dealing with BERT derivatives~\cite{wehnert2022applying, rogers2020primer}. Since those parameters don’t always produce the best results and their use may still result in a model remaining undertrained, an alternative strategy was adopted for choosing the upper limit of training epochs that tracks validation loss and terminates training only when the conditions are met. 
	
	\subsubsection{Legal text classification}
	The samples within the experimental dataset were collected from two main portals. The first data set was collected from the Scientific Judicial Portal(SJP)\footnote{https://sjp.moj.gov.sa/}, which is run by the of Ministry of Justice in Saudi Arabia. SJP is the largest specialized information database in the field of the judiciary in the Kingdom of Saudi Arabia. It is the ideal solution for specialists in the judicial and legal field, including judges, lawyers, trained lawyers, academics, prosecutors, graduate students, and others. The second data we collected was from the Board of Grievances(BoG) portal\footnote{https://www.bog.gov.sa/en/ScientificContent/Pages/default.aspx}, which is as they stated in their website: \textit{"The Board tested a judiciary academic series in the name of (judge library) and its distribution among the Board judges (hard copy and soft copy) to increase cognitive formation with them, a state which its effect shall be reflected on the judiciary verdicts they issue, including academic references in administrative, commercial and penal judiciary formed of 32 volumes in addition to judiciary verdicts"}.
	
	Since the existing documents in both datasets can belong to several categories depending on the submission details, they are suitable for the task of long  length legal document classification. Three different classes of documents were selected from the SJP dataset and ten classes from the BoG dataset. Since practically all documents from certain classes are headlined in the same way, for the classification task it was necessary to truncate the parts of the document containing easily identifiable indicators of the class. Due to this omission, the model needs to analyze the entire content instead of simply deriving the conclusion based on the first few lines. This modification was implemented with all classes.
	
	\subsubsection{Keyword Extraction}
	
	Unfortunately, compared to English and some other Latin languages, there is no ready-made and well-prepared data for research purposes, especially in the field of natural languages processing of legal texts in Arabic. Therefore, we built our data in this task with the help of professionals in the Arab legal field. This dataset consists of approximately 8K legal documents containing main keywords, extracted manually by these professionals. We did a preprocessing and cleaning of the data and extracted about 37640 sentences containing both keywords and others. The average sentence length was no more than 65 words, as we did a sentence segmentation process so that the sentence does not lose its meaning or break. We tagged the keywords in the sentence with the number 1 and the others with the number 0.
	
	\subsubsection{Named Entity Recognition}
\begin{figure*}[!ht]
		\centering
		\caption{The main NER tags}
		\label{fig:table2}
		\includegraphics[width=0.4\linewidth]{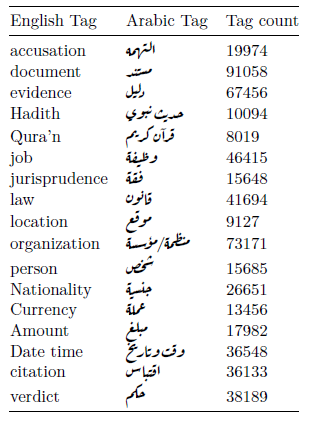}
	\end{figure*}

	This dataset also was created inside the research department at Elm company. It contains more than 311K sentences including thousands of distinct entities of seventeen different sequence tags that labeled manually by multiple human annotators as of part of our CourtNLP project in Elm research\footnote{https://www.elm.sa/en/research-and-innovation/Pages/Research.aspx}. All the used classes are shown in Figure~\ref{fig:table2}. The main objective of NER procedure is to associate a label belonging to a particular class to each included word. It’s important to note that some complex named entities could stretch across multiple words, but are always contained in a single sentence. The predominant sentence representation form used in this field is IOB, with words that start a name of an entity marked with B, internally located words marked as I, and other tokens marked with O.
	
	\begin{table*}[!ht]
		\caption{The overall results of all fine-tuned models in the legal text classification task on BoG Dataset}
		\label{tabBoG}
		\centering
		\begin{tabular}{@{}llll@{}}
			\toprule
			Model / Macro-Average & Precision & Recall      & F1-score       \\ \midrule
			Arabertv2-Large~\cite{antoun2020arabert}         & 0.850387  & 0.810795  & 0.827078 \\
			ARBERT~\cite{abdul2020arbert}   & 0.802514  & 0.821973  & 0.812820\\
			mBERT~\cite{devlin2018bert} & 0.702017  & 0.635928 & 0.598267 \\
			AraLegal-BERT (base)    & \textbf{0.89276}  & \textbf{0.89173} & \textbf{0.89098} \\ \bottomrule
		\end{tabular}
	\end{table*}
	
	\begin{table*}[!ht]
		\caption{The overall results of all fine-tuned models in the legal text classification task on SJP Dataset}
		\label{tabSJP}
		\centering
		\begin{tabular}{@{}llll@{}}
			\toprule
			Model / Macro-Average & Precision & Recall      & F1-score       \\ \midrule
			Arabertv2-Large~\cite{antoun2020arabert}  & 0.885678  & 0.886816  & 0.884516\\
			ARBERT~\cite{abdul2020arbert} & 0.837714  & 0.834804 & 0.843827 \\
			mBERT~\cite{devlin2018bert} & 0.814763  & 0.780226 & 0.782501 \\
			AraLegal-BERT (base)     & \textbf{0.92395}  & \textbf{0.92133} & \textbf{0.92210} \\ \bottomrule
		\end{tabular}
	\end{table*}
	
	\section{Results}
	\subsection{Impact of pre-training}
	We trained two models from scratch, the first one is a base model that contains 12 layers and the second one is a large model consists of 24 layers, as in the original BERT. As could be anticipated, full-sized with 24-layer model that was trained from scratch has a much better ability to fulfill the pre-training objectives than the base model with 12-layer. However, after the pre-training stage is completed the base model displays similar level of loss to original BERT that is trained on general datasets. In particular, it is faster to adjust to narrowly defined niches, while can be a great advantage for domain-focused applications such legal. For this reason, the content of the training set plays into the selection of the suitable training method. We did not experiment on the large model yet, and all the fine-tuning results were based on the base model since we find that it gives much better accuracy than the general Arabic BERT models in the three defined NLU tasks.  
	
	\subsection{Results discussion}
	We have divided the datasets into training, validation and testing datasets with respect to all of the three tasks. Here we will discuss the testing results. The evaluation was conducted using standardized hyperparameters such as batch size and sequence length, with several different datasets suitable for legal text classification, keyword extraction, and named entity recognition. 
	
	In the first task of legal text classification, the best option in this task is determined based on experimental results, for example, in this way we found that multiple strategies can be used to get around BERT’s limit to sequence length of 512 tokens, but the ‘head \& tails’ strategy where only the first 128 and the last 382 tokens are kept performs the best such as the work in~\cite{sun2019fine}. Tables~\ref{tabBoG} and~\ref{tabSJP} summarize the over all results of our model compared to three BERT variations for Arabic language on the classification task with two datasets namely, SJP and BoG. On the dataset of BoG, AraLegal-BERT, outperformed the all models by 0.7\% in F1-Macro average higher than the highest model of the three, which is ARABERT-v2large. Similarly, our model also outperformed the rest of the models in the SJP dataset, with a good difference between it and ArabBERTv2-large, which is approximately 0.4\% in F1-Macro average.
	
	On the other hand, as for the tasks related to the named enity recognition as well as the keyword extraction, we followed the same procedure that was done in our previous work~\cite{al2021arabic}, taking into account that no new layers were added to the model, a linear layer was used to make the word and sequence-level tagging. In these two tasks, the results are more different, and there is a significant difference between the performance of AraLegal-BERT and the rest of the models, where AraLegal-BERT outperformed ARABERT-v2large~\cite{antoun2020arabert} with a difference of almost 21\% in the F1-Macro average in the task of extracting named entities as depects in Table~\ref{tabNER}, while the difference is much greater in the keyword extraction task, where AraLegal-BERT outperformed the highest model which is ARBERT~\cite{abdul2020arbert} with a significant difference equal to almost 26\% in the F1-Macro average as shown in Table~\ref{tabKw}. 
	
	We denote that the general BERT models were not only less in the F1-macro score, but also in the recall-macro score, where they are not able to retrieve most of the required words compared to AraLegal-BERT. Finally, it is worth noting that the used AraLegal-BERT model is the base version; nevertheless, it has outperformed the rest of the models in all the three defined tasks with less memory size and faster performance as well as good accuracy score.

	\begin{table*}[!ht]
		\caption{The overall results of all fine-tuned models in the Keywords Extraction Task}
		\label{tabKw}
		\centering
		\begin{tabular}{@{}llll@{}}
			\toprule
			Model / Macro-Average & Precision & Recall      & F1-score       \\ \midrule
			Arabertv2-Large~\cite{antoun2020arabert} & 0.765979 & 0.470098    & 0.582625 \\
			ARBERT~\cite{abdul2020arbert}  & 0.789460 & 0.524715    & 0.630420  \\
			mBERT~\cite{devlin2018bert}& 0.721356 & 0.375075   & 0.493533 \\
			AraLegal-BERT (base)    & \textbf{0.93481}  & \textbf{0.84449} & \textbf{0.88736} \\ \bottomrule
		\end{tabular}
	\end{table*}
	
	\begin{table*}[!ht]
		\caption{The overall results of all fine-tuned models in the Named Entity Recognition Task}
		\label{tabNER}
		\centering
		\begin{tabular}{@{}llll@{}}
			\toprule
			Model / Macro-Average & Precision & Recall      & F1-score       \\ \midrule
			Arabertv2-Large~\cite{antoun2020arabert}         & 0.891934  & 0.450073  & 0.598261 \\
			ARBERT~\cite{abdul2020arbert} & 0.889916  & 0.413266 & 0.564423 \\
			mBERT~\cite{devlin2018bert} & 0.886825  & 0.326475 & 0.477254 \\
			AraLegal-BERT (base)     & \textbf{0.89848}  & \textbf{0.73644} & \textbf{0.80943} \\ \bottomrule
		\end{tabular}
	\end{table*}
	
	\section{Conclusions and future work}
	Based on the findings of our experiments, the pre-training of BERT for a specific domain is better than the general language models in certain NLU task. For this reason, we present AraLegal-BERT model that were trained exclusively for Arabic legal texts and are able to make very accurate predictions on three main NLU tasks, namely Legal text classification, named entity recognition and keywords extraction. Crucially, the level of difficulty of the task is correlated with the gains from choosing the right training strategy, as the importance of domain-specific vocabulary and semantics becomes more pronounced. The tested version of AraLegal-BERT model is the base, cost-efficient version suitable for a broad range of Arabic legal textual applications. In the near future, future work will focus on additional possibilities for pre-training other models such as Electra and Roberta models for several NLU tasks in the Arabic legal domain with small, base and large versions. 
	
	\section*{Acknowledgements}
	This work was supported and funded by Elm Company Research department.

	\bibliography{anthology,custom}
	\bibliographystyle{acl_natbib}
	
\end{document}